\documentclass[lettersize,journal]{IEEEtran}
\usepackage{amsmath,amsfonts}
\usepackage{algorithmic}
\usepackage{algorithm}
\usepackage{array}
\usepackage[caption=false,font=normalsize,labelfont=sf,textfont=sf]{subfig}
\usepackage{textcomp}
\usepackage{stfloats}
\usepackage{url}
\usepackage{verbatim}
\usepackage{graphicx}
\usepackage{cite}
\begin{document}

\title{Exploring Depth Information for Detecting Manipulated Face Videos}


\DeclareRobustCommand*{\IEEEauthorrefmark}[1]{%
    \raisebox{0pt}[0pt][0pt]{\textsuperscript{\footnotesize\ensuremath{#1}}}}
    
\author{
	\IEEEauthorblockN{
		Haoyue Wang\IEEEauthorrefmark{1}, 
		Sheng Li\IEEEauthorrefmark{1},~\IEEEmembership{Member,~IEEE,}
        Ji He\IEEEauthorrefmark{2}, 
		Zhenxing Qian\IEEEauthorrefmark{1},~\IEEEmembership{Senior Member,~IEEE,} 
		Xinpeng Zhang\IEEEauthorrefmark{1},~\IEEEmembership{Senior Member,~IEEE,}
		and Shaolin Fan\IEEEauthorrefmark{2}} 
        
	\IEEEauthorblockA{\IEEEauthorrefmark{1}Fudan University, Shanghai, China}
    
	\IEEEauthorblockA{\IEEEauthorrefmark{2}Shanghai Information Security Testing Evaluation and Certification Center, Shanghai, China}
}

\maketitle

\begin{abstract}
Face manipulation detection has been receiving a lot of attention for the reliability and security of the face images/videos. Recent studies focus on using auxiliary information or prior knowledge to capture robust manipulation traces, which are shown to be promising. As one of the important face features, the face depth map, which has shown to be effective in other areas such as face recognition or face detection, is unfortunately paid little attention to in literature for face manipulation detection. In this paper, we explore the possibility of incorporating the face depth map as auxiliary information for robust face manipulation detection. To this end, we first propose a Face Depth Map Transformer (FDMT) to estimate the face depth map patch by patch from an RGB face image, which is able to capture the local depth anomaly created due to manipulation. The estimated face depth map is then considered as auxiliary information to be integrated with the backbone features using a Multi-head Depth Attention (MDA) mechanism that is newly designed. We also propose an RGB-Depth Inconsistency Attention (RDIA) module to effectively capture the inter-frame inconsistency for multi-frame input. Various experiments demonstrate the advantage of our proposed method for face manipulation detection.
\end{abstract}


\section{Introduction}
The development of deep learning techniques has made face manipulation an easy task. People can manipulate face images/videos using a variety of deepfake schemes \cite{thies2019deferred, thies2016face2face, perov2020deepfacelab, perez2003poisson, li2019faceshifter, huang2020fakepolisher, Kowalski2018faceswap}. Manipulated faces are usually difficult to be distinguished by human eyes, which seriously challenges the reliability and security of face images/videos. It is of paramount importance to develop advanced and accurate face manipulation detection schemes.

Researchers have devoted a lot of effort to the task of face manipulation detection. Various deep neural networks (DNN) are proposed to spot the difference between real and manipulated face images, such as ResNet \cite{he2016resnet}, Xception \cite{rossler2019faceforensics++}, MesoNet \cite{afchar2018mesonet}, and EfficientNet \cite{tan2019efficientnet}. Recently, researchers start to explore different auxiliary information or prior knowledge to facilitate face manipulation detection, including the blending boundary \cite{li2020face}, guided residuals\cite{10017352}, identity \cite{cozzolino2021id}, pre-generated face attention mask \cite{zi2020wilddeepfake, Schwarcz2021finding}, face information in the frequency domain \cite{gu2022exploiting, chen2021local}, and the face texture features \cite{zhu2021face,zhao2021multi,10202582}. Such a strategy is shown to be promising for performance boosting. When the input is a set of face video frames, researchers are keen to exploit the inconsistency of the facial features among the fake face frames, such as irregular eye blinking \cite{liy2018exposingaicreated}, lip motions \cite{Haliassos2021lips} or some specific feature representations \cite{sohrawardi2019poster,nguyen2019capsule,zheng2021exploring,gu2021spatiotemporal,zhang2021detecting,10.1145/3503161.3547913,gu2022delving,9795231}.

These schemes achieve good performance when the face manipulation schemes are known in training. In real world applications, however, the detection model is often faced with fake faces generated by unknown manipulation schemes. The classifiers learnt from the detailed manipulation traces in one dataset may not be robust against those from another dataset, which results in severe performance reduction in cross-database scenarios. More efforts are needed to discover and learn more robust face manipulation features to improve the generalization ability of face manipulation detection models. 

In this paper, we explore the possibility to estimate and incorporate the depth map as auxiliary information for robust face manipulation detection. The rationales behind this: 
\begin{figure*}
    \centering
    \includegraphics[width=0.98\textwidth]{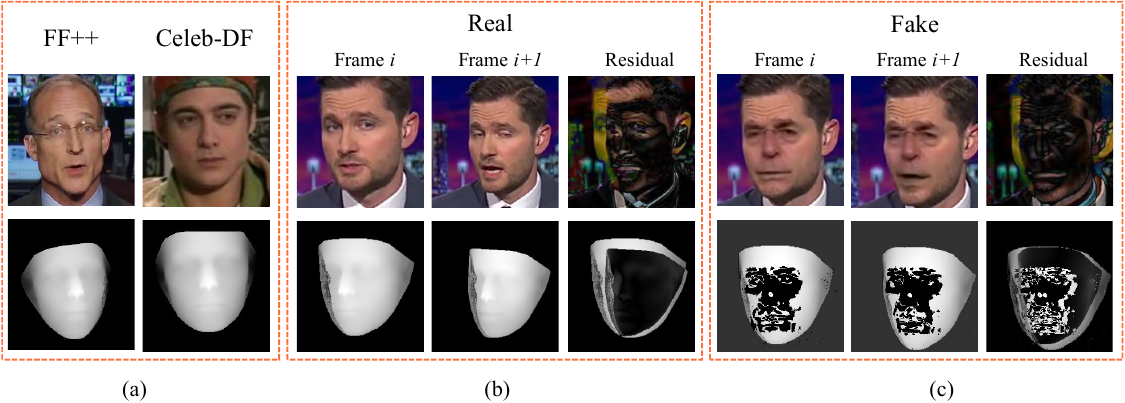}
    \caption{Examples of the face depth maps and residuals. (a) Depth maps of real face images from different sources (FF++\cite{rossler2019faceforensics++} and Celeb-DF \cite{li2020celebdf}); (b) depth maps of two consecutive real face frames and the corresponding residual; (c) depth maps of two consecutive fake face frames and the corresponding residual. Please refer to Section \ref{fdmt} for the computation of the ground truth face depth maps.}
    \label{fig:residual_intuition}
\end{figure*}

\begin{itemize}
    \item The face depth map tends to be stable among the face images that are collected from different sources (see Fig.\ref{fig:residual_intuition} (a)), while the manipulation would distort the face depth maps. Take the popular generative DNN based face manipulation as an example, the fake face region will either have no depth (if it is computer-generated) or have abnormal depth features around the boundary (if it is swapped from another real face image). 
    \item The face movement in a fake video will result in abnormal changes of the depth feature. The residual of the face depth map between two consecutive frames tends to be more discriminative in capturing the manipulation traces than directly computing the residual of the two frames in the RGB space. As shown in Fig.\ref{fig:residual_intuition} (b) and (c), the face residuals in the RGB space appear similarly for real and fake face videos. In the depth space, however, abrupt changes could be observed in the manipulated area in the residual computed from the fake face videos.   
\end{itemize}

Now the question becomes how we could accurately estimate the depth from a real or fake face image. There have been studies to estimate the depth map from a two-dimensional RGB face image \cite{feng2018joint, jin2021face, wu2021dual, kang2022facial}, however, all these schemes assume the face image is real and captured from a human face which could be considered as a physical object with a relatively smooth surface. The estimated face depth maps are globally smooth, which are not sensitive to face manipulation operations. To deal with this issue, we propose a Face Depth Map Transformer (FDMT) for face depth estimation, which is capable of capturing the local patch-wise depth variations due to face manipulation. Next, we propose a Multi-head Depth Attention (MDA) mechanism to effectively incorporate the face depth map into the backbones for face manipulation detection. We further design an RGB-Depth Inconsistency Attention (RDIA) module to well support video-level face manipulation detection, where the correlation of the spatial-temporal inconsistency between the RGB and depth space is measured and incorporated to effectively capture the inter-frame inconsistency of the fake face videos. 

The proposed method is shown to be significantly better than the existing schemes in the cross-database scenario, which also achieves good performance in the intra-database scenario for detecting fake faces. We evaluate the generalization ability of the proposed method on three popular face manipulation detection backbones including Xception \cite{chollet2017xception}, ResNet50 \cite{he2016resnet}, and EfficientNet \cite{tan2019efficientnet}, all demonstrate the effectiveness of the proposed method for face manipulation detection. The contributions of this paper are summarized below.

\begin{itemize}
    \item We explore the possibility of using the face depth map, which is seldom considered in the area of face manipulation detection, for performance boosting.
    \item We propose a Face Depth Map Transformer (FDMT) for generating local patch-wise depth features that are sensitive to face manipulation.    
    \item We propose a Multi-head Depth Attention (MDA) mechanism to effectively integrate our face depth maps into different backbones for face manipulation detection. 
    \item We propose an RGB-Depth Inconsistency Attention (RDIA) module to measure the correlation of the spatial-temporal inconsistency between the RGB and depth space of the face, so as to further boost the performance for video-level face manipulation detection.
 \end{itemize}

\section{Related Works}

In this section, we first briefly introduce techniques for face manipulation and then review recent works on face manipulation detection and face depth map estimation.

\subsection{Face Manipulation}
Recent face manipulation techniques have been divided into three categories based on different forgery purposes: Face Swap, Face Attribute Editing, and Expression Replacement. Face Swap refers to replacing the face of one person in an image or video with another person's face to achieve identity forgery. The common and publicly available techniques for this are Deepfake\cite{perov2020deepfacelab} and FaceSwap\cite{Kowalski2018faceswap}. Face Attribute Editing is mostly done by modifying facial attributes such as color, skin, age, etc. using GANs like StarGAN\cite{choi2018stargan} and AttGAN\cite{he2019attgan}. Expression Replacement involves falsifying facial expressions without changing the identity of the person. Typical methods for this include Face2Face\cite{thies2016face2face} and NeuralTextures\cite{thies2019deferred}.

\subsection{Face Manipulation Detection}
\textbf{Image-Based Methods:}~With the continuous development of deep learning, various DNN backbones have been proposed for face manipulation detection. Wang \textit{et al.} \cite{wang2020cnn} apply ResNet \cite{he2016resnet} to classify real or fake face images. Rossler \textit{et al.} \cite{rossler2019faceforensics++} use Xception \cite{chollet2017xception} as a baseline DNN model, which can achieve satisfactory performance on intra-database evaluations. Afchar \textit{et al.} \cite{afchar2018mesonet} design a compact network MesoNet for video based face manipulation detection. Zhao \textit{et al.} \cite{zhao2021multi} propose to take advantage of the EfficientNet \cite{tan2019efficientnet} for face manipulation detection, which can achieve comparable performance to Xception. There are also patch-based face manipulation detection approaches proposed to extract the subtle manipulated traces located in the image patches. Chai \textit{et al.} \cite{chai2020makes} take advantage of a patch-based classifier with limited receptive fields in the image. The works in \cite{chen2021local, zhao2021learning} further consider the patch similarity to facilitate the classification, where each patch is equally treated and processed during the patch feature learning. Zhang \textit{et al.} \cite{zhang2022patch} propose a Patch Diffusion (PD) module to fully exploit the patch discrepancy for effective feature learning.

A lot of recent studies focus on exploring effective auxiliary information or prior knowledge for the task of face manipulation detection, which are shown to be promising. Li \textit{et al.} \cite{li2020face} take the facial blending boundary as an indicator for the existence of manipulation. Dang \textit{et al.} \cite{dang2020detection} propose to incorporate the position of the face manipulation area to make the network focus on the manipulation traces. Zi \textit{et al.} \cite{zi2020wilddeepfake} extract and fuse the face mask and organ mask into an attention mask to make the network pay attention to the fake area. Schwarcz \textit{et al.} \cite{Schwarcz2021finding} generate masks of the important parts of the face image to perform multi-part detection. The works in \cite{zhu2021face,zhao2021multi} introduce different approaches to extract the face texture features to guide the network for better detection of the manipulation cues. Masi \textit{et al.} \cite{masi2020two} adopt a dual-branch network structure, one of which is a fixed filter bank to extract the face feature in the frequency domain for auxiliary information. Similarly, the works in \cite{frank2020leveraging,qian2020thinking, gu2022exploiting, chen2021local} propose different approaches to treat the frequency domain face information as auxiliary information to boost the performance. 

\begin{figure*}[htbp]
    \centering
    \includegraphics[width=0.98\textwidth]{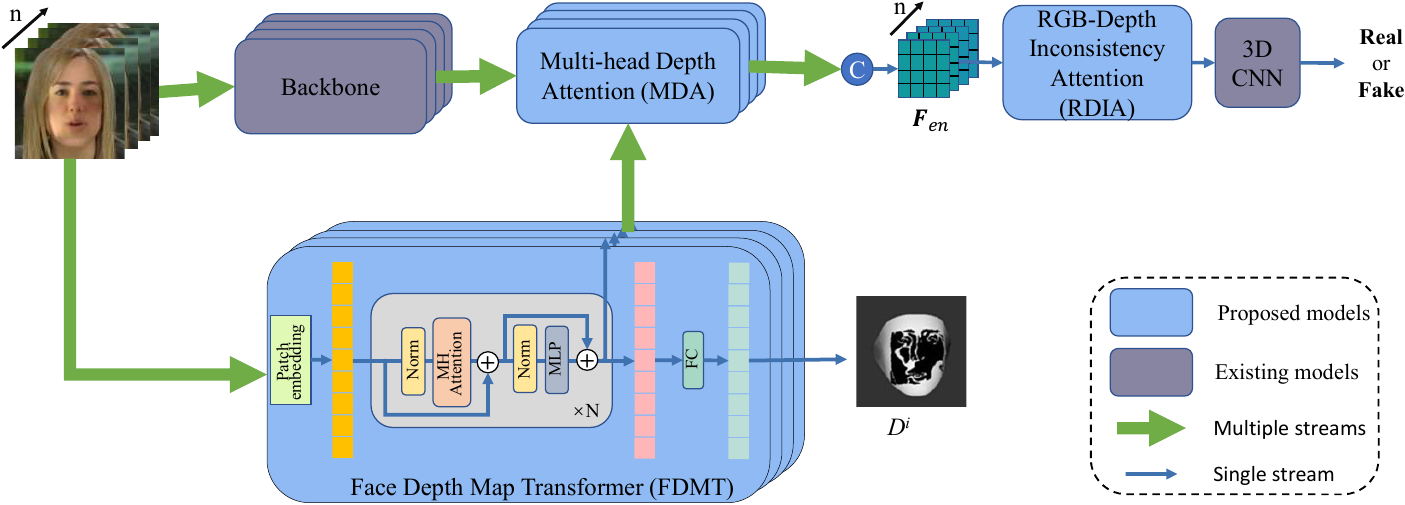}
    \caption{An overview of the proposed method for face manipulation detection.}
    \label{fig:framework}
\end{figure*}

\textbf{Video-Based Methods:}There are also schemes proposed specifically for video-level face manipulation detection. Most of these schemes focus on extracting discriminative features to represent the inconsistency among the manipulated face video frames. Zheng \textit{et al.} \cite{zheng2021exploring} construct a temporal transformer by combining the features of different face frames into time series for input. Gu \textit{et al.} \cite{gu2021spatiotemporal} observe that the motion between adjacent frames in real videos is more smooth than fake ones, where a Spatial-Temporal Inconsistency Learning (STIL) scheme is proposed to capture the inconsistency. Zhang \textit{et al.} \cite{zhang2021detecting} propose a Temporal Dropout 3-dimensional Convolutional Neural Network (TD3DCNN) to detect the temporal incoherence of the face frames. A spatial-temporal dropout transformer is proposed in \cite{10.1145/3503161.3547913} to make full use of the spatial-temporal inconsistency of local facial areas among different frames. Cozzolino \textit{et al.} \cite{cozzolino2021id} proposes to use the identity information for detecting manipulated face videos, which requires a real face video for reference to conduct the detection.


\subsection{Face Depth Map Estimation}
In general, face depth map estimation tries to construct the depth information from one or more two-dimensional RGB face images. Feng \textit{et al.} \cite{feng2018joint} propose a Position map Regression Network (PRNet) to exquisitely estimate the face depth. Since the scene depth is relatively easier to obtain compared with the face depth map, Jin \textit{et al.} \cite{jin2021face} apply scene depth knowledge for face depth map estimation. Wu \textit{et al.} \cite{wu2021dual} propose a depth uncertainty module to learn a face depth distribution instead of a fixed depth value. Kang \textit{et al.} \cite{kang2022facial} propose a StereoDPNet to perform depth map estimation from dual-pixel face images.

The face depth map is shown to be a good feature for face related machine learning tasks. Chiu \textit{et al.}  \cite{chiu2021high} develop a segmentation-aware depth estimation network, DepthNet, to estimate depth maps from RGB face images for accurate face detection. Wang \textit{et al.} \cite{wang2020deep} argue that the depth map can reflect discriminative clues between live and spoofed faces, where the PRNet \cite{feng2018joint} is adopted to estimate the face depth map for face anti-spoofing. Zheng \textit{et al.} \cite{zheng2021attention} also takes the face depth map as auxiliary information for face anti-spoofing, where a symmetry loss is proposed for reliable face depth estimation. 

Motivated by the effectiveness of the face depth map in the aforementioned applications, we believe it is worthy of investigation to see how we could take advantage of such information for face manipulation detection. We think the face depth map is a robust feature against different capturing devices, which could be helpful when we are encountering face images collected from unknown sources. To this end, we propose a Face Depth Map Transformer (FDMT) to estimate the face depth map from both the original and manipulated face images. This is then treated as auxiliary information to be fused with the backbone feature by a Multi-head Depth Attention (MDA) mechanism newly designed for performance boosting. We also propose an RGB-Depth Inconsistency Attention (RDIA) module to effectively learn the inconsistency among different fake face frames.

\section{The Proposed Method}
\subsection{Overview}

Fig.\ref{fig:framework} gives an overview of our proposed method for estimating and integrating the face depth map for face manipulation detection. Given a set of $n$ sequential face frames as input, we partition each frame into a set of non-overlapping patches for patch-wise face depth estimation, where we propose a Face Depth Map Transformer (FDMT) to construct the face depth patch by patch. Next, we propose a Multi-head Depth Attention (MDA) to integrate the depth feature extracted from the FDMT (before the fully connected layer) with the backbone feature to enhance the feature representation of each frame. We further propose an RGB-Depth Inconsistency Attention (RDIA) to enhance the feature representation for capturing the inter-frame inconsistency. The enhanced features are then fed to the the rest of network for classification.

\subsection{Face Depth Map Transformer (FDMT)}
\label{fdmt}
\begin{figure}
    \centering
    \includegraphics[width=0.35\textwidth]{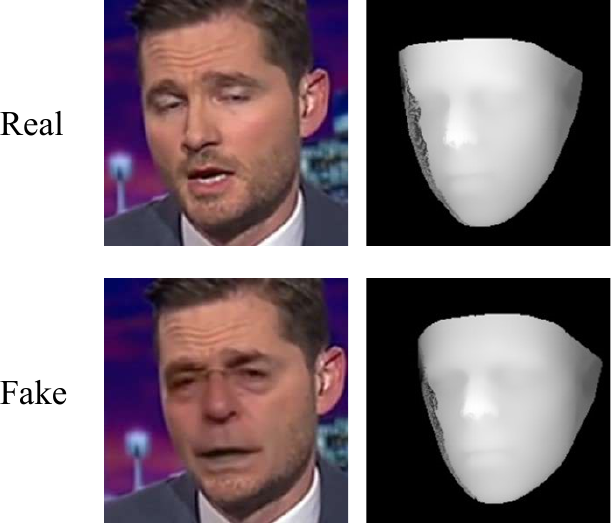}
    \caption{Examples of the estimated face depth map using PRNet \cite{feng2018joint}.  Images in the ``Real” row are real face image and depth map. Images in the ``Fake” row are manipulated face image and depth map. The face images are selected from FF++ \cite{rossler2019faceforensics++}.
    \label{fig:generatedepth}}
\end{figure}

Face manipulation usually alters the original face image locally to change the face appearance. Such an operation will cause abrupt changes in the face depth map, which is unfortunately difficult to be captured by using the existing face depth map estimation schemes. Because they assume the face image is captured from a human face with a relatively smooth surface, as shown in Fig.\ref{fig:generatedepth}. To make the face depth map sensitive to face manipulation, we propose here a Face Depth Map Transformer (FDMT) to estimate the face depth features patch by patch. Our FDMT is supervised by a set of ground truth face depth maps which are generated specifically for face manipulation detection. Next, we elaborate in detail on how we generate the ground truth as well as the FDMT. 

\begin{figure}
    \centering
    \includegraphics[width=0.35\textwidth]{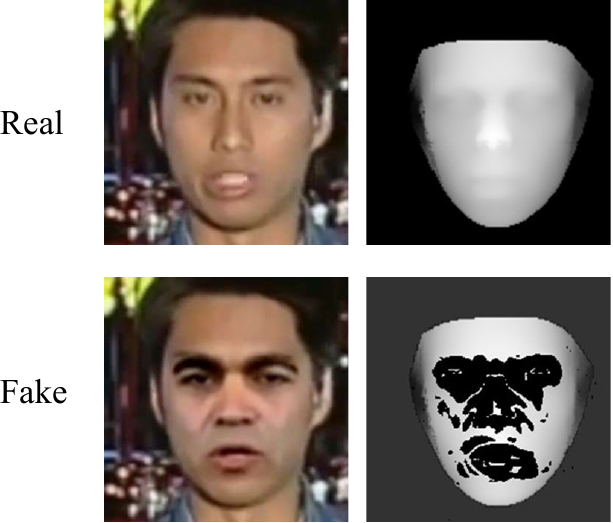}
    \caption{Examples of the ground truth face depth map. Images in the ``Real'' row are the real face image and the ground truth. Images in the ``Fake'' row are manipulated face image and the ground truth. The face images are selected from FF++ \cite{rossler2019faceforensics++}. \label{fig:depthmap}}
\end{figure}

The ground truth face depth map should properly reflect the depth of the real face, fake face, and background regions of a manipulated face image. As suggested by \cite{gu2022exploiting}, the fake face region can be obtained by binarizing the residual between a manipulated face image and its original version. And there are pre-trained models available for depth estimation from real face images. These offer us the possibility to generate appropriate face depth maps to be served as ground truth for face manipulation detection. 

Given a face image, we first generate its face depth map based on a pre-trained face depth map estimation model (PRNet\cite{feng2018joint}). The PRNet automatically segments the image into the face region and background region, where the depth of the background region is set as 0 and the depth within the face region is represented as non-zero positive integers, and a larger value means closer to the camera. Let's denote the output of PRNet as $D(x,y)$ for the pixel located at $(x,y)$ in the face image. The ground truth face depth at pixel $(x,y)$ is computed as 
\begin{equation}
     G(x,y) = \begin{cases}
                0,   &  if (x,y) \in \emph{fake face region} \\
                \Omega(D(x,y) + \lambda) ,& \emph{otherwise},
            \end{cases}
\end{equation}
where $\Omega$ is the operation to prevent overflow (i.e., set the values larger than 255 as 255) and $\lambda$ is a positive integer. As such, we have well-separated depth values for different regions, where the depth of the fake face and background regions are set as 0 and $\lambda$, the depth value of the real face region is within the range from $\lambda$ to $255$. The ground truth face depth of each image patch is then computed as the average depth value within this patch based on $G(x,y)$. Fig.\ref{fig:depthmap} gives examples of our ground truth face depth map for face manipulation detection.

The structure of our FDMT is similar to ViT \cite{dosovitskiy2020vit}. We divide an input face image into a set of $P$ non-overlapping patches with positions embedded. Then, the position embedded patches are processed into $T$ transformer blocks, where each block contains two normalization layers, a multi-head attention unit, and a multi-layer perceptron (MLP). The output of the last transformer block is fed into a fully connected layer to produce a $P$ dimensional vector representing the depth value of each patch.  

\subsection{Multi-head Depth Attention (MDA)}
\begin{figure*}
    \centering
    \includegraphics[width=0.85\textwidth]{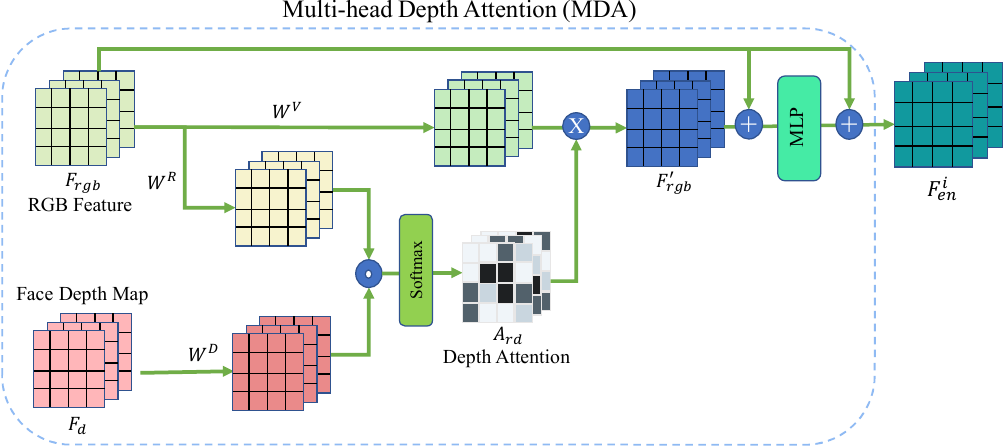}
    \caption{The network structure of Multi-head Depth Attention.}
    \label{fig:MDA}
\end{figure*}

With the face depth map available, the next question is how to effectively integrate it into the backbone. A straight forward way is to concatenate it with the backbone features extracted from the RGB face frame for enhancement (termed as the RGB feature for simplicity). We could also directly use it as attention to weight the RGB feature. However, both strategies ignore the correspondence between the RGB feature and the face depth map, whose correlations are not fully exploited during the integration. Here, we propose to jointly learn a depth attention by taking both the RGB feature and face depth map into consideration. Please refer to Fig.\ref{fig:MDA} for details of the network structure of the Multi-head Depth Attention.

Given the RGB feature $F_{rgb}$ extracted from the backbone and the face depth feature $F_{d}$ extracted from FDMT, where $F_{rgb}$ and $F_{d}$ are with the same height and width. The RGB feature contains the color and texture information, as well as the manipulation clues in the spatial feature spaces. While the face depth feature offers the corresponding depth anomalies caused by manipulation as auxiliary information. To take advantage of such correspondences, we measure the similarity between the RGB feature and the face depth feature using the dot product below
\begin{equation}
     F_{rd} = (F_{d}W^{D}) \cdot (F_{rgb}W^{R}),
     \label{eq:DP_RGB}
\end{equation}
where $W^{R}$ and $W^{D}$ are the trainable weight matrices for the RGB feature and the face depth feature, ``$\cdot$'' is the dot product operation, and the biases are omitted for simplicity. Then, we use softmax to convert the similarity $F_{rd}$ into a depth attention by
\begin{equation}
     A_{rd} = softmax(\frac{F_{rd}}{\sqrt{d}}),
\end{equation}
where $d$ is a scaling factor equivalent to the number of channels in the RGB feature. The depth attention is eventually used to enhance the RGB feature by
\begin{equation}
     F'_{rgb} = A_{rd} \odot (F_{rgb}W^{V}),
\end{equation}
where $W^{V}$ is a trainable weight matrix, ``$\odot$'' is the element-wise multiply operation. The backbone feature is then enhanced below by fusing the RGB features before and after the depth attention 
\begin{equation}
    F_{en} = F_{rgb} + \text{MLP}(F'_{rgb}+F_{rgb}).
\end{equation}

Next, we adopt the multi-head strategy \cite{Vaswani2017attention} on our depth attention to achieve an $l$-head depth attention, which is given as 
\begin{multline}
\label{eq:multihead}
    \text{MultiHead}(F_{d}W^{D}, F_{rgb}W^{R}, F_{rgb}W^{V}) = \\
        \text{Concat}(d_{1},d_{2},\dots,d_{l})W^{O},
\end{multline} 
where $d_i$ refers to the $i$-th head with input being $F_{d}W^{D}$, $F_{rgb}W^{R}$, $F_{rgb}W^{V}$ and $F_{en}$ being the output, ``Concat'' is the concatenation operation, and $W^{O}$ is a weight matrix to aggregate the outputs of different heads. Note that the trainable matrices for $F_{d}$ and $F_{rgb}$ are not shared among different heads.  

\begin{figure*}
    \centering
    \includegraphics[width=0.98\textwidth]{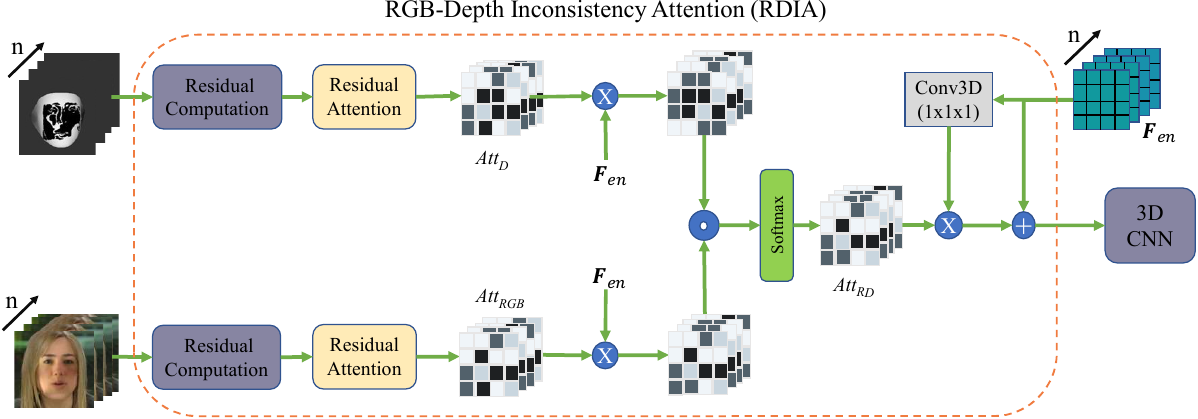}
    \caption{The network structure of RGB-Depth Inconsistency Attention.}
    \label{fig:RDIA}
\end{figure*} 

\subsection{RGB-Depth Inconsistency Attention (RDIA)}

\begin{figure*}
    \centering
    \includegraphics[width=0.85\textwidth]{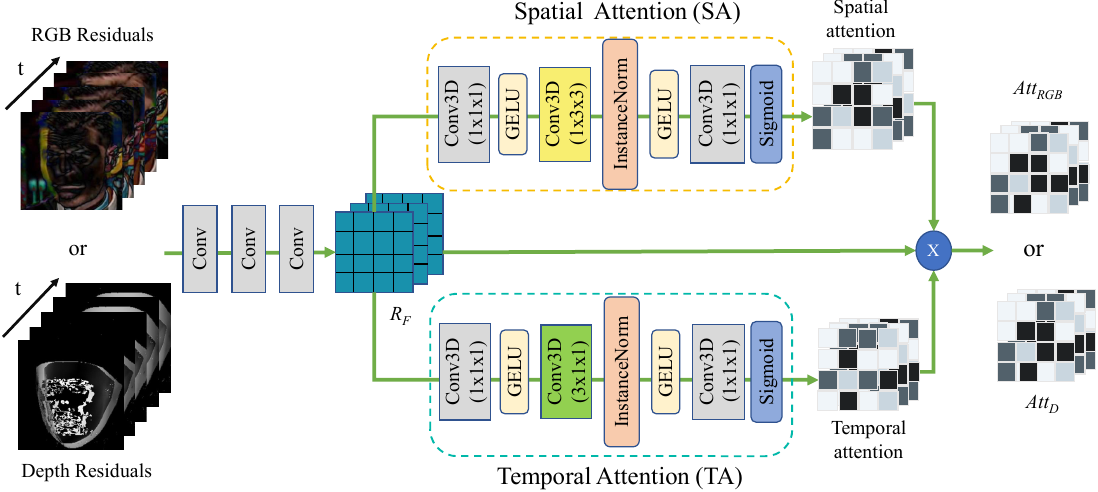}
    \caption{The network structure of the residual attention.}
    \label{fig:resattention}
\end{figure*}

Let's denote the $n$ input face frames as $\boldsymbol{I} = \{I_{i}\}^{n}_{i=1}$ with $\boldsymbol{D} =\{D^i\}^{n}_{i=1}$ being the depth maps estimated by FDMT, we denote the face features after MDA as $\boldsymbol{F}_{en} = \{F^i_{en}\}^{n}_{i=1}$. The purpose of the RDIA is to further enhance the face features by learning the inconsistency among different frames in the depth and RGB space. Please refer to Fig.\ref{fig:RDIA} for details of the network structure of the RGB-Depth Inconsistency Attention. The main component in the RDIA is a residual attention (RA) module which combines both the spatial and temporal attention from the residuals of the frames in the depth and RGB space, as shown in Fig.\ref{fig:resattention}. Next, we explain in detail regarding how the RA works in the depth space. The input of RA is a set of $n-1$ depth residuals $\boldsymbol{R}_{d} = \{r^{i}_{d}\}^{n-1}_{i=1}$,where $r^{i}_{d} = D^{i+1}-D^i$. We use a set of convolutional layers to preprocess the residuals by 
\begin{equation}
\label{eq:res_process}
         R_{F} = \chi (\boldsymbol{R}_{d}),
\end{equation}
where $\chi(\cdot)$ is a network consists of three 2D convolutional layers. The residual attention for the depth map is then computed by
\begin{equation}
\label{eq:Attention_r}
        Att_{D} = R_{F} \odot \text{softmax}(\text{SA}(R_{F}) \odot \text{softmax}(\text{TA}((R_{F})), 
\end{equation}
where ``$\odot$'' is the element-wise multiply operation, ``$\text{SA}$'' computes the spatial attention, ``$\text{TA}$'' obtains the temporal attention. The $\text{SA}$ and $\text{TA}$ differ mainly in the kernels used for 3D convolution. Specifically, the TA uses a $3\times1\times1$ 3D convolution kernel to perform convolution operations only along the temporal direction. While the $\text{SA}$ adopts a kernel sized $1\times3\times3$ to conduct the convolution only on the spatial domain. Please refer to Fig.\ref{fig:resattention} for details of the network structure of the $\text{SA}$ and $\text{TA}$.

For the same token, we can also compute a piece of residual attention for the frames $\boldsymbol{I}$ in the RGB space (say $Att_{RGB}$). By considering both the residual attentions in the depth space and the RGB space, we obtain a RGB-Depth inconsistency attention by 
\begin{equation}
        Att_{RD} = \text{softmax}(\phi(Att_{RGB} \odot \boldsymbol{F}_{en})) \cdot \psi(Att_{D} \odot \boldsymbol{F}_{en})),
\end{equation}
where $\phi(\cdot)$ and $\psi(\cdot)$ are 3D convolution operations with kernel size of $1\times1\times1$, ``$\cdot$'' is the dot product operation. The RGB-Depth inconsistency attention is eventually used to enhance the face features by
\begin{equation}
        F' = Att_{RD} \odot \upsilon(\boldsymbol{F}_{en}) + \boldsymbol{F}_{en},
\end{equation}
where $\upsilon(\cdot)$ refers to a $1\times1\times1$ convolution. The enhanced features are then fed to a 3DCNN for classification.

\subsection{Loss Function}
We adopt the SSIM loss and the MSE loss to evaluate the similarity between the estimated and the ground truth face depth map. The SSIM loss is formulated as 
\begin{equation}
    \mathcal{L}_{ssim} = \frac{(2\mu_{a}\mu_{b}+c_{1})(2\sigma_{ab}+c_{2})}{(\mu_{a}^{2}+\mu_{b}^{2}+c_{1})(\sigma_{a}^{2}+\sigma_{b}^{2}+c_{2})},
\end{equation}
where $\mu_{a}$, $\mu_b$ represent the mean of the estimated and ground truth face depth map; $\sigma_{a}^{2}$, $\sigma_{b}^{2}$ denote the corresponding variance; $\sigma_{ab}$ is the covariance; $c_{1}$ and $c_{2}$ are small positive integers to avoid the division by zero. The MSE loss is given by \begin{equation}
    \mathcal{L}_{patch\_mse} = \sum_{i=1}^{M}\sum_{p=1}^{P}{||a_{i,p}-b_{i,p}||_{2}},
\end{equation}
where $a_{i,p}$ and $b_{i, p}$ are the depth values of the $p$-th patch in the estimated and ground truth face depth map for the $i$-th training sample, respectively.

The total loss is computed as

\begin{equation}
    \label{eq:loss}
    \mathcal{L}_{total}=\mathcal{L}_{c} + \alpha\cdot\mathcal{L}_{ssim} + \beta\cdot\mathcal{L}_{patch\_mse},
\end{equation}
where $\mathcal{L}_{c}$ is the backbone loss for image classification, $\alpha$ and $\beta$ are the weights to balance different loss terms.

\section{Experiments}

\subsection{Setup}
\textbf{Dataset:}~We use two large-scale face manipulation datasets: FaceForensics++ (FF++) \cite{rossler2019faceforensics++}, Celeb-DF \cite{li2020celebdf} for experiments. The FF++ dataset contains 1,000 real videos, with 720 videos for training, 140 videos for validation and 140 videos for testing. Each video has four versions using different manipulation methods, which are DeepFakes (DF) \cite{Tora2018deepfakes}, Face2Face (F2F) \cite{thies2016face2face}, FaceSwap (FS) \cite{Kowalski2018faceswap} and Neural Textures (NT) \cite{thies2019deferred}. Besides, each video has three compression levels, which are RAW, High Quality (c23), and Low Quality (c40). The Celeb-DF dataset includes 590 raw videos collected from YouTube and 5639 corresponding face manipulated videos, which cover refined manipulated face videos from different genders, ages, and races with similar quality to those transmitted in real world scenarios.

\textbf{Evaluation Metrics:}~We use the detection accuracy (ACC) and Area under the Curve (AUC) for evaluation, which are two common indicators in literature for evaluating face manipulation detection schemes. 

\textbf{Implementation Details:}~We take Xception \cite{chollet2017xception} as a backbone by default to evaluate the performance of our proposed method, where we integrate our Multi-head Depth Attention between the seventh to the eighth block of the Xception. The length of the input video frames is $n=8$. For each frame, we follow the suggestion given in \cite{rossler2019faceforensics++} to automatically and conservatively crop the facial area into a square, which is then resized to $224 \times 224$ for training and testing. To obtain the fake face region mask, we first calculate the pixel-level difference between the manipulated frame and its corresponding original frame. Then, we binarize the pixel-level difference with a threshold of $15$. The mask is used to generate the ground truth face depth map with $\lambda=50$, which is then normalized to $[0,1]$ for training. Our FDMT contains $T=12$ blocks with eight attention heads for each block, where the input image is partitioned into $P=14\times14$ patches. Both the values of $\alpha$ and $\beta$ are set as $0.7$ for the total loss. The model is trained with Adam optimizer \cite{Kingma2015AdamAM} with a learning rate $3\times10^{-4}$ and a weight decay $10^{-4}$. We train our model on two RTX 3090 GPUs with a batch size of $4$.

\subsection{Comparisons}

\begin{table} 
    \caption{The AUC (\%) of different schemes on the FF++ for intra-database evaluation.}
    \label{tab:FF++}
    \centering
    \resizebox{0.50\columnwidth}{!}{
        \begin{tabular}{cccccc}
        \hline
        Method               & FF++     \\ 
        \hline
        Meso4 \cite{afchar2018mesonet} & 84.70 \\
        Xception  \cite{rossler2019faceforensics++} &96.74 \\
        Two-branch \cite{masi2020two} &  93.18  \\
        X-Ray \cite{li2020face} & 92.80  \\
        $F^{3}Net$ \cite{qian2020thinking} &98.10 \\
        ADDNet-3d \cite{zi2020wilddeepfake}      &98.30 \\
        LTW \cite{sun2021domain}   &98.50 \\
        Multi-attention \cite{zhao2021multi} &\textbf{99.80} \\
        FT-two-stream\cite{9408664} &  92.47   \\
        SPSL   \cite{liu2021spatial} &96.91   \\
        DIANet \cite{hu2021dynamic} &90.40  \\
        FInfer \cite{hu2022finfer} &95.67\\
        STDT \cite{10.1145/3503161.3547913}   & \textbf{99.80} \\
        Ours (Xception)-\uppercase\expandafter{\romannumeral1}       &97.98 \\ 
        Ours (Xception)-\uppercase\expandafter{\romannumeral2}       &98.40 \\ 
        \hline
        \end{tabular}
        }
\end{table}

\textbf{Intra-database Evaluation:}~We train our model on the whole training set of FF++(c23) and test it on FF++(c23) for intra-database evaluation. Table.\ref{tab:FF++} gives the AUC of face manipulation detection among different schemes on FF++, where the performance of the existing schemes are all duplicated from literature. Here, ``Ours (Xception)-\uppercase\expandafter{\romannumeral1}'' refers to the case that we only use one frame for input without using the RDIA module (i.e., image-level detection), while ``Ours (Xception)-\uppercase\expandafter{\romannumeral2}'' means video-level detection. It can be seen that our proposed method achieves satisfactory performance with over 98\% AUC. Compared with using the original Xception, our proposed method increases the AUC by 0.63\% for image-level detection and 1.66\% for video-level detection.  

\begin{table}
    \caption{The AUC(\%) of different schemes for cross-database evaluation.}
    \label{tab:cross-database}
    \centering
    \resizebox{0.73\columnwidth}{!}{
        \begin{tabular}{cccc}
        \hline
        Method         & FF++ DF           & Celeb-DF  \\ 
        \hline
        Meso4 \cite{afchar2018mesonet}          & -         &54.80  \\
        Xception  \cite{rossler2019faceforensics++}  &94.55 &73.27  \\
        $F^{3}Net$ \cite{qian2020thinking}        &-        &65.17  \\
        Two-branch \cite{masi2020two}           &-          &73.41  \\
        ADDNet-3d  \cite{zi2020wilddeepfake}    &96.22      &60.85  \\
        FT-two-stream\cite{9408664}             &-          &65.56  \\
        LTW \cite{sun2021domain}                &-          &64.10  \\
        Multi-attention \cite{zhao2021multi}    &-          &67.44  \\
        DIANet  \cite{hu2021dynamic}            &90.40      &70.40  \\
        DSANet  \cite{wu2021dsanet}             &96.88      &73.71  \\
        SPSL   \cite{liu2021spatial}            &-          &76.88  \\
        STIL \cite{gu2021spatiotemporal}        &97.12      &75.58       \\
        FInfer \cite{hu2022finfer}              &           &70.60 \\
        STDT \cite{10.1145/3503161.3547913}           &-          &69.78       \\
        LDIL   \cite{gu2022delving}             &98.19      &77.65          \\
        Ours (Xception)-\uppercase\expandafter{\romannumeral1}  &97.31    &80.58 \\
        Ours (Xception)-\uppercase\expandafter{\romannumeral2}   & \textbf{98.33}   & \textbf{83.35}  \\ 
        \hline
        \end{tabular}
        }
\end{table}

\textbf{Cross-database Evaluation:}~By following the suggestion given in most of the existing works, we train our model on the whole training set of FF++ (c23) dataset and test it on the Celeb-DF test set for cross-dataset evaluation. Table.\ref{tab:cross-database} shows the AUC of different schemes, where the performance on the DF subset of FF++(c23) is also given for reference. It can be seen from Table.\ref{tab:cross-database} that our method achieves the best in both cases for video-level detection. It is worth noting that our method is significantly better than the existing schemes for cross-dataset evaluation, with over 5.7\% higher AUC when tested on Celeb-DF.

\begin{table}
    \caption{The performance of integrating the MDA into different blocks in Xception.}
    \label{tab:insertion}
    \centering
    \resizebox{0.72\columnwidth}{!}{
    \begin{tabular}{ccc}
    \hline
    Different blocks for integration & ACC        \\ 
    \hline
     -           & 93.22        \\
    B3    & 95.23       \\
    B7    & \textbf{96.30}          \\
    B11  &   94.39 \\ 
    \hline
    \end{tabular}
    }
\end{table}

\subsection{Sensitivity of Multi-head Depth Attention (MDA)}
As shown in Fig.\ref{fig:framework}, our proposed MDA has to be integrated into the backbone for feature enhancement. In this section, we test the sensitivity of our MDA on the F2F subset of FF++ (c40) by integrating it with the feature maps of different blocks in the default Xception backbone. We denote the 13 blocks in Xception as B1, B2, ..., B13 from input to output. We integrate the MDA with the feature maps extracted from three different blocks: B3, B7, and B11, the results of which are given in Table.\ref{tab:insertion}. It can be seen that all the integration methods improve the detection accuracy and the integration with the features extracted from the center block (i.e., B7) achieves the best.

\subsection{Generalization Ability against Different Backbones}
In this section, we integrate our proposed method with three popular face manipulation detection backbones including Xception \cite{chollet2017xception}, Resnet50 \cite{he2016resnet}, and EfficientNet \cite{tan2019efficientnet}, where the EfficientNet is the runner up in the Facebook Deepfake Detection Challenge \cite{dolhansky2019deepfake}. We train all the models on the whole training set of FF++, and test them on Celeb-DF. We conduct the MDA integration with the feature maps extracted right after the seventh block, the third layer, and the twelfth block in the Xception, Resnet50, and EfficientNet. Here, ``Ours ($*$)-\uppercase\expandafter{\romannumeral1}'' and ``Ours ($*$)-\uppercase\expandafter{\romannumeral2}'' refer to the image-level and video-level detection, respectively.

Table.\ref{tab:backbone} gives the performance before and after using our proposed method for integration. It can be seen that, regardless of the backbones, our proposed method is able to boost the performance for the cross-database scenario. At the image level, the performance gains for Xception, ResNet50, and EfficientNet are $7.31\%$, $1.66\%$, and $3.41\%$ in terms of AUC, respectively. At the video level, the corresponding performance gains are $10.08\%$, $3.52\%$, and $4.23\%$ respectively. These indicate the good generalization ability of our proposed method for performance boosting on different backbones.

\begin{table}
\centering
\caption{The AUC(\%) of cross-database evaluation for different backbones before and after integrating our proposed method.}
\label{tab:backbone}
\resizebox{0.60\columnwidth}{!}{
    \begin{tabular}{ccc}
        \hline
        Backbone  & Celeb-DF \\  
        \hline 
        Xception\cite{chollet2017xception}          &73.27         \\
        Ours (Xception)-\uppercase\expandafter{\romannumeral1}   & 80.58 \\
        Ours (Xception)-\uppercase\expandafter{\romannumeral2}   &\textbf{83.35} \\
        \hline
        ResNet50\cite{he2016resnet}         &65.66          \\
        Ours (ResNet50)-\uppercase\expandafter{\romannumeral1}    &67.32  \\
        Ours (ResNet50)-\uppercase\expandafter{\romannumeral2}  &\textbf{69.18}\\
        \hline
        EfficientNet\cite{tan2019efficientnet}    & 72.07          \\
        Ours (EfficientNet)-\uppercase\expandafter{\romannumeral1}  &75.48  \\ 
        Ours (EfficientNet)-\uppercase\expandafter{\romannumeral2}  & \textbf{76.30} \\
        \hline
    \end{tabular}
    }
\end{table}

\begin{figure}
    \centering
    \includegraphics[width=0.45\textwidth]{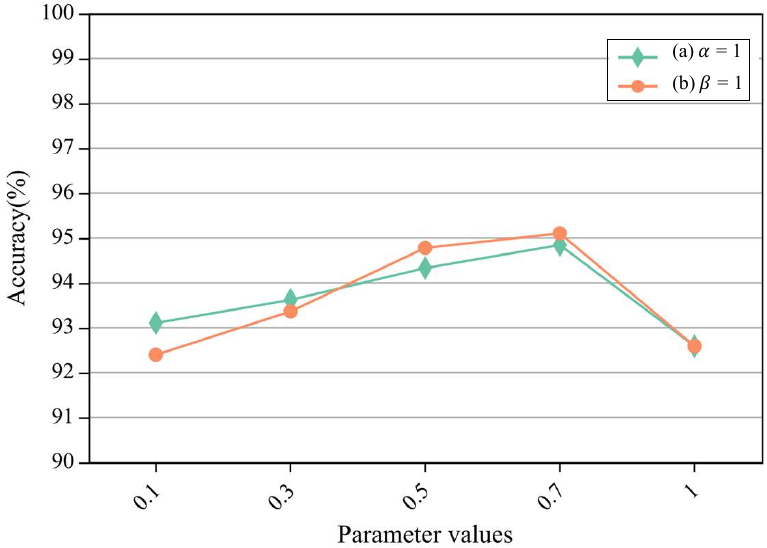}
    \caption{The detection performances achieve by (a) varying $\beta$ when $\alpha$ is fixed as 1 and (b) varying $\alpha$ when $\beta$ is fixed as 1.}
    \label{fig:hyper_parameters}
\end{figure}

\subsection{Parameter Setting}~As indicated by the loss function in Eq.(13), the weight parameters $\alpha$ and $\beta$ may affect the final combination of the losses. To study the influence of these two hyper-parameters, we conducted experimental analysis by fixing one parameter and adjusting the other, where the models are trained and tested on the DF subset in FF++. Fig.\ref{fig:hyper_parameters} illustrates the variation of accuracy under different settings. The results indicate that as the values of $\alpha$ or $\beta$ increase, the accuracy improves significantly. The highest accuracy is achieved when $\alpha$ or $\beta$ equals 0.7, with accuracies of 94.85\% and 95.11\%, respectively. However, when both $\alpha$ and $\beta$ are set to 1, the weight of the depth map estimation task is too high, resulting in a negative impact on the backbone classification task.

\subsection{Ablation Study}
To verify the effectiveness of our proposed Face Depth Map Transformer (FDMT), Multi-head Depth Attention (MDA), and RGB-Depth Inconsistency Attention (RDIA), we evaluate them separately in this section, where all the models are trained and tested on the DF subset in FF++ (c40) with the backbone fixed as Xception, and we take the backbone feature extracted from the seventh block in Xception for fusion or attention when necessary.

\begin{table}
    \caption{Ablation study for the FDMT.}
    \label{tab:Ab_depth}
    \centering
    \resizebox{0.55\columnwidth}{!}{
    \begin{tabular}{cccc}
        \hline
        Depth Map & Fusion & ACC  \\ 
        \hline
        -         & -      & 93.22  \\
        PRNet\cite{feng2018joint}    & concat &93.87  \\
        FDMT      & concat & \textbf{95.16}\\
        \hline
    \end{tabular}
    }
\end{table}

\textbf{Effectiveness of FDMT:}~To demonstrate the effectiveness of FDMT, we conduct two additional experiments here: 1) we concatenate the face depth features extracted from FDMT with the backbone features, 2) we concatenate the face depth features extracted from an existing face depth estimation model PRNet \cite{feng2018joint} with the backbone features. The concatenated features are then passed through a 1x1 convolutional layer to obtain fused features for the subsequent Xception blocks. Table.\ref{tab:Ab_depth} gives the ACC of the aforementioned experiments as well as those of the Xception backbone. It can be seen that both two experiments achieve higher ACC compared with using the Xception backbone only. This again indicates that the face depth map is indeed helpful for the face manipulation detection task. By simply concatenating an existing face depth map with the backbone feature, we are able to achieve 0.65\% improvement in ACC. While the face depth map estimated using our proposed FDMT is more effective for performance boosting, with 1.94\% of improvement in ACC compared with the Xception backbone.

\begin{figure*}
    \centering
    \includegraphics[width=0.95\textwidth]{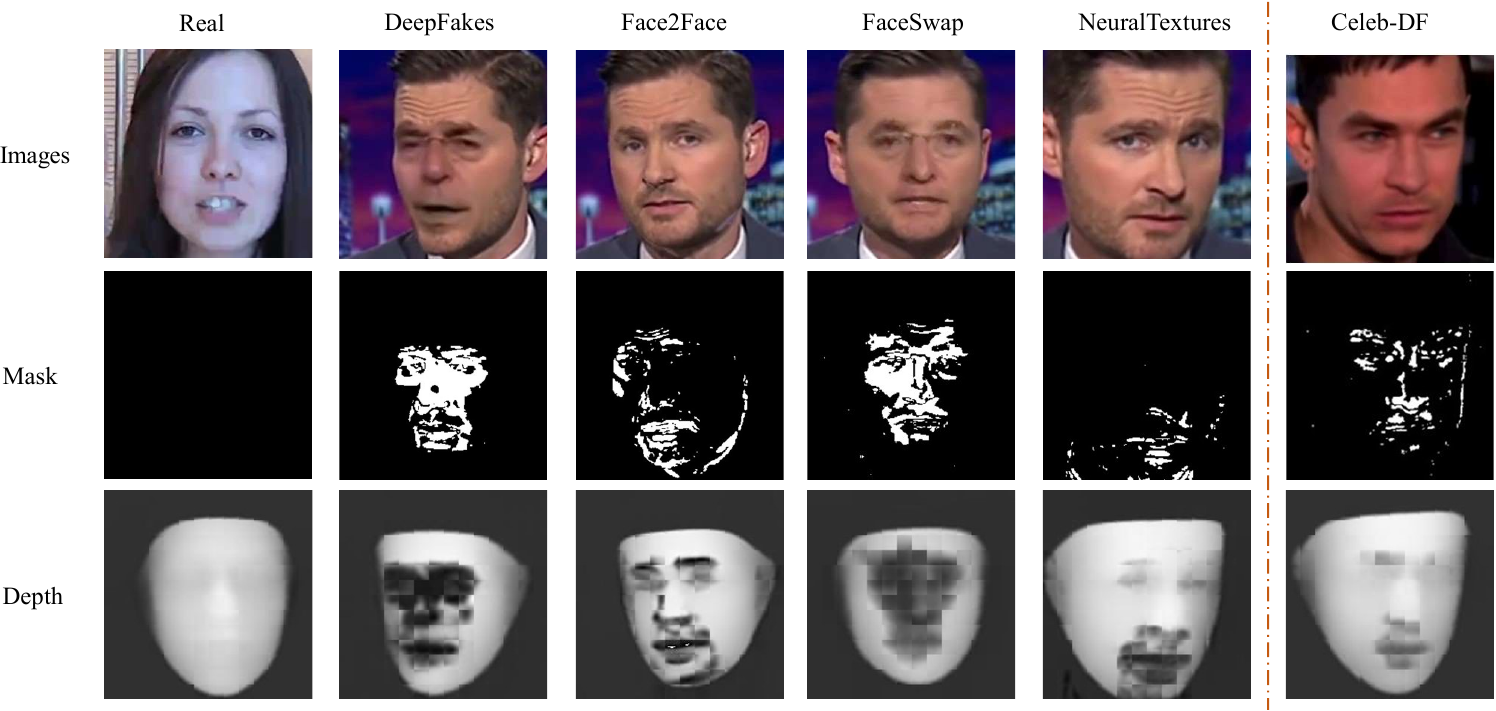}
    \caption{Examples of the face depth maps estimated using our proposed FDMT. The first row shows the face images selected from different databases, the second row gives the masks of the fake face region (shown in white) and the third row presents the estimated face depth maps. The first five columns give examples from different subsets in FF++ \cite{rossler2019faceforensics++}, and the sixth column illustrates an example from Celeb-DF \cite{li2020celebdf}.}
    \label{fig:depth_generate}
\end{figure*}

Fig.\ref{fig:depth_generate} gives some examples of the estimated face depth map from real and fake face images using the proposed FDMT. The images on the first three columns are from the FF++ dataset, the image on the fourth column is selected from Celeb-DF. It can be seen that our FDMT can effectively estimate the face depth, which is able to capture the anomaly in face depth caused by face manipulation for face images from different sources. This further demonstrates the ability of our FDMT to extract robust face depth features for face manipulation detection.

\begin{table}
    \caption{Ablation study for the MDA.}
    \label{tab:DepthAttention}
    \centering
    \resizebox{0.60\columnwidth}{!}{
        \begin{tabular}{cccc}
        \hline
        Depth Map & Attention & ACC \\ \hline
        -         & -         &  93.22   \\
        -     & MSA           &93.62  \\
        PRNet\cite{feng2018joint}     & MDA & 94.33  \\
        FDMT      & MDA       & \textbf{96.30}\\ \hline
        \end{tabular}
    }
\end{table}

\textbf{Effectiveness of MDA:}~To verify the effectiveness of our proposed MDA, we conduct two more experiments here: 1) we replace our FDMT with the PRNet for face depth estimation, 2) we incorporate the popular multi-head self-attention (MSA) \cite{Vaswani2017attention} with the backbone feature without using the face depth map. Table.\ref{tab:DepthAttention} gives the ACC of face manipulation detection for different models. It can be seen that, by simply using the MSA, we are able to achieve 0.4\% higher in ACC compared with using the Xception backbone only. Our MDA is able to further increase the ACC which are around 0.7\% and 2.7\% higher than that of using the MSA by depth attention based on PRNet and our proposed FDMT, respectively. Compare with the results of directly concatenating the face depth map (see Table.\ref{tab:Ab_depth}), our MDA achieves more performance gain with around 1\% improvement in ACC for both the face depth maps estimated using the PRNet and our proposed FDMT. These results indicate that the multi-head based attention mechanism works for the task of face manipulation detection, and our MDA is superior to the existing MSA with the help of the face depth map. 

\begin{table}
    \caption{Ablation study for the RDIA.}
    \label{tab:RDIA}
    \centering
    \resizebox{0.72\columnwidth}{!}{
        \begin{tabular}{cccc}
        \hline
        Inconsistency Learning Scemes & ACC \\ \hline
         3DCNN           &94.66 \\
         RDIA  &  \textbf{96.30}\\ \hline
        \end{tabular}
    }
\end{table}

\textbf{Effectiveness of RDIA:}~To verify the effectiveness of our proposed RDIA, we conduct video-level detection by using 3DCNN to learn the inconsistency among different video frames instead of using RDIA. Table.\ref{tab:RDIA} gives the performance of using 3DCNN and RDIA for video-level detection, where the features of each frame are extracted using FDMT and MDA. It can be seen that the ACC is improved by 1.64\% using our RDIA instead of using 3DCNN. This indicates that our RDIA works better in capturing the inconsistency cues among different fake video frames. 

\begin{figure*}[t]
    \centering
    \includegraphics[width=0.95\textwidth]{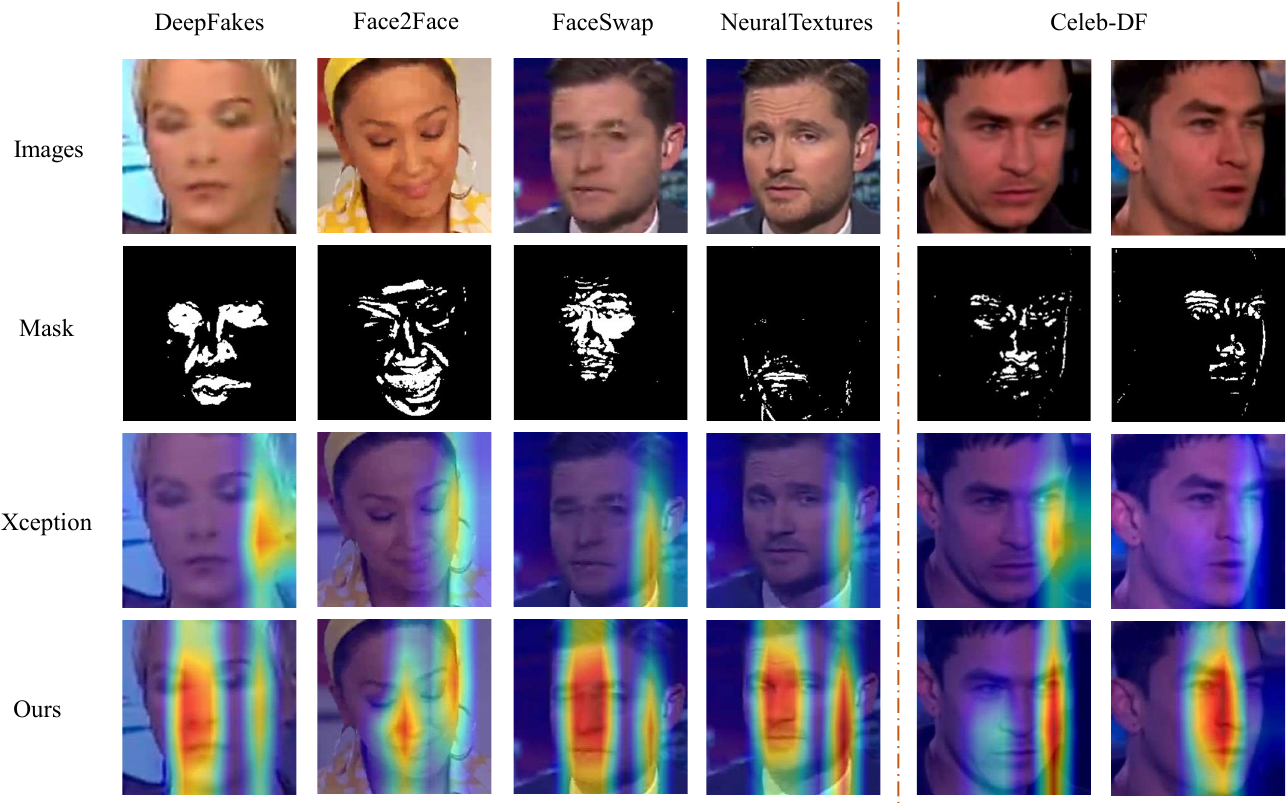}
    \caption{Visulization of the features before and after using our proposed method on the Xception backbone. From top to bottom: fake face images (first row), the masks of the fake face region (second row), and visualization of the features before (third row) and after (fourth row) using our proposed method. The first four columns give examples from different subsets in FF++ \cite{rossler2019faceforensics++}, and the fifth and sixth column illustrates an example from Celeb-DF \cite{li2020celebdf}.} 
    \label{fig:GradCam}
\end{figure*}

Next, we visualize the Gradient-weighted Class Activation Mapping (Grad-CAM) \cite{selvaraju2017grad} of the backbone features before and after the integration of our proposed method, as shown in Fig.\ref{fig:GradCam}. It can be seen that, by using our proposed method, the backbone feature is able to focus more on the fake face region, which is helpful for accurate face manipulation detection.

\section{Conclusion}
In this paper, we explore the possibility of using the face depth map to facilitate face manipulation detection. To extract representative face depth maps from the manipulated face frames, we design a Face Depth Map Transformer (FDMT) to estimate the face depth maps patch by patch, which is effective in capturing the local depth anomaly created due to the manipulation. To appropriately integrate the face depth feature, we further propose a Multi-head Depth Attention (MDA) mechanism to enhance the backbone features via a depth attention which is computed by the scale dot product between the face depth feature and the backbone feature of the RGB face image. To well facilitate video-level detection, we propose an RGB-Depth Inconsistency Attention (RDIA) module to effectively capture the inconsistency among different frames in the RGB and depth spaces. Experimental results indicate that our proposed method is particularly helpful in the cross-database scenario with over 5.7\% higher AUC than the existing schemes.  

\bibliographystyle{IEEEtran}
\bibliography{reference}

\vfill

\end{document}